# Research on Detection of Floating Objects in River and Lake Based on AI Intelligent Image Recognition

Jingyu Zhang[1,a], Ao Xiang[2,b], Yu Cheng[3,c], Qin Yang[4,d], Liyang Wang[5,e]

[1]The University of Chicago, The Division of the Physical Sciences, Analytics, Chicago, IL, USA

[2]University of Electronic Science and Technology of China, School of Computer Science & Engineering (School of Cybersecurity), Digital Media Technology, Chengdu, Sichuan, China

[3]Columbia University, The Fu Foundation School of Engineering and Applied Science, Operations Research, New York, NY, USA

[4]University of Electronic Science and Technology of China, School of Integrated Circuit Science and Engineering (Exemplary School of Microelectronics), Microelectronics Science and Engineering, Chengdu, Sichuan, China

[5]Washington University in St. Louis, Olin Business School, Finance, St. Louis, MO

[a]simonajue@gmail.com, [b]xiangao1434964935@gmail.com, [c]yucheng576@gmail.com, [d]yqin0709@gmail.com, [e]liyang.wang@wustl.edu

**Abstract:** With the rapid advancement of artificial intelligence technology, AI-enabled image recognition has emerged as a potent tool for addressing challenges in traditional environmental monitoring. This study focuses on the detection of floating objects in river and lake environments, exploring an innovative approach based on deep learning. By intricately analyzing the technical pathways for detecting static and dynamic features and considering the characteristics of river and lake debris, a comprehensive image acquisition and processing workflow has been developed. The study highlights the application and performance comparison of three mainstream deep learning models – SSD, Faster-RCNN, and YOLOv5 – in debris identification. Additionally, a detection system for floating objects has been designed and implemented, encompassing both hardware platform construction and software framework development. Through rigorous experimental validation, the proposed system has demonstrated its ability to significantly enhance the accuracy and efficiency of debris detection, thus offering a new technological avenue for water quality monitoring in rivers and lakes.

## 1.Introduction

In the contemporary era where environmental conservation is increasingly becoming a global focal point, the cleanliness and preservation of rivers and lakes are particularly paramount. Among the primary factors affecting water quality, the presence of floating debris stands out, thereby rendering its monitoring and management exceedingly urgent. Nonetheless, conventional monitoring methods not only consume significant time and effort but also exhibit suboptimal efficiency, thus failing to meet the rigorous standards demanded by current environmental protection norms. In light of these circumstances, this study introduces AI-powered image recognition technology, aiming to explore a more efficient and precise approach to automatic detection of floating debris. Through a comprehensive examination of AI's applications in image recognition, this research successfully pioneers a novel model for detecting debris in rivers and lakes, optimizing monitoring procedures and substantially enhancing the speed and accuracy of data processing. This groundbreaking advancement foreshadows a future wherein the management and preservation of river and lake environments will be more scientifically and efficiently executed, thereby contributing to the realization of sustainable development objectives.

**2. AI Intelligent Image Recognition Technology**

*2.1. Basic Principle*

The advancement of AI-driven image recognition technology hinges upon the transformation of image data into machine-interpretable information, enabling the recognition and classification of targets. Its fundamental principles encompass three pivotal steps: image preprocessing, feature extraction, and classification decision-making. Image preprocessing aims to enhance the quality of image data through tasks such as denoising and contrast enhancement, facilitating subsequent processing. In the feature extraction phase, algorithms analyze processed images to identify key features conducive to classification, such as shapes, colors, or textures. Ultimately, in the classification decision-making step, leveraging machine learning models like deep learning neural networks, objects within the images are accurately recognized and classified based on the extracted features. The success of this process heavily relies on algorithmic design and the quality of training data, ensuring the accuracy and reliability of recognition outcomes [1].

*2.2. Feature Detection*

*2.2.1. Static Feature Detection*

In the realm of image processing techniques, static feature detection stands as a fundamental and pivotal task, wielding decisive influence over the comprehension and subsequent analysis of images. Static features, as implied by their name, are attributes extracted from static images that remain invariant over time, furnishing crucial foundations for image recognition and classification.

Among static features, color features stand out as the most direct and commonly utilized. They analyze the color values of individual pixels within an image, unveiling its visual perceptual information. By scrutinizing the distribution of colors within an image, one can effectively differentiate between its various components or identify distinct image entities. Below unfolds the expression formula for color moments, wherein Formula (1) represents the first moment; Formula (2) denotes the second moment; Formula (3) encapsulates the third moment. Here, pi,j signifies the ith color component of the jth pixel, while N signifies the total number of pixels within the image [2].

$$\mu_i = \frac{1}{N}\sum_{j=1}^{N} p_{i,j} \qquad (1)$$

$$\sigma_i = \left(\frac{1}{N}\sum_{j=1}^{N}(p_{i,j} - \mu_i)^2\right)^{1/2} \qquad (2)$$

$$s_i = \left(\frac{1}{N}\sum_{j=1}^{N}(p_{i,j} - \mu_i)^3\right)^{1/3} \qquad (3)$$

Texture features delineate the surface characteristics within an image, encompassing attributes such as smoothness, roughness, and directional patterns of the regions depicted. These attributes are instrumental in discerning materials depicted within images or in parsing scenes, as distinct objects and surfaces often exhibit unique textural motifs.

Corner features, on the other hand, are focused on identifying the corners present within an image, denoting points where abrupt changes in edges occur. They play a crucial role in tasks such as contour detection and shape recognition owing to their unique positions and attributes, serving as vital connectors between different image features.

Spatial relationship features encompass the relative positions and arrangement patterns among various objects or elements within an image. Understanding such features is paramount for grasping the overall structure of an image and the relationships among objects within a scene. They aid in deciphering spatial hierarchies and depth cues within an image, offering clues for its three-dimensional interpretation.

In summary, the detection and analysis of these static features furnish a robust foundation for image recognition and comprehension. Advancements in technology have led to more precise and efficient extraction and utilization of these features, significantly propelling progress within the realm of computer vision [3].

*2.2.2. Dynamic Feature Detection*

In the realm of video streaming or sequential image sequences, dynamic feature detection endeavors to identify and track objects in motion. Unlike static image analysis, dynamic feature analysis emphasizes the extraction of temporal variations, pivotal for comprehending video content, monitoring scenarios, or implementing advanced driver assistance systems, among other applications. The crux of dynamic feature detection lies in accurately capturing and analyzing the motion states of objects within image sequences.

Background subtraction is a prevalent technique for dynamic feature extraction, identifying moving objects by comparing the current frame with a background model. Its advantage lies in its ability to distinguish foreground moving objects from static backgrounds, albeit its sensitivity to object detection in dynamic backgrounds and susceptibility to lighting variations.

Optical flow relies on the analysis of pixel motion speed and direction across image sequences to infer object motion within a scene. By computing pixel displacement between consecutive frames, optical flow yields detailed insights into object velocity and movement direction. While suitable for analyzing intricate motions and estimating changes in object speed within a scene, this method incurs relatively high computational costs [4].

Frame differencing, on the other hand, detects moving objects by analyzing differences between consecutive frames or multiple frames. This method, being straightforward and efficient, promptly responds to changes in object motion, making it suitable for real-time video analysis applications. However, frame differencing may struggle with blur issues arising from swiftly moving objects and may exhibit suboptimal performance in detecting objects with minor movement. The workflow of frame differencing is illustrated in Figure 1.

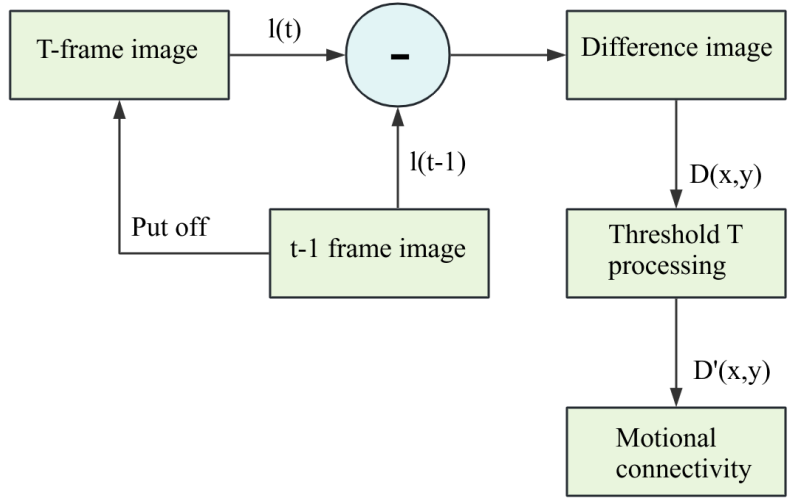

**Figure 1.** Flowchart of the interframe difference method

The steps implemented through inter-frame differencing are as follows:

1. Denote the video image sequences' respective frames at t and t-1 as I(t) and I(t-1) respectively. The corresponding pixel points of the two frames are denoted as I(t)(xy) and I(t-1)(x,y). Finally, compute the difference between the two frames to obtain D(x, y), as detailed in formula (4).

$$D(x, y) = |I(t)(x, y) - I(t-1)(x, y)| \quad (4)$$

2. According to set the threshold T and binarize the difference image to get the binarized image as shown in formula (5).

$$D'(x, y) = \begin{cases} 255, & if \quad D(x, y) > T \\ 0, & others \quad D(x, y) \leq T \end{cases} \quad (5)$$

3. The final moving target image can be obtained by concatenating the moving regions in the binarized image D(x, y).

In conclusion, the development of dynamic feature detection techniques aims to improve the accuracy and efficiency of moving object detection. Various methods have their own advantages and disadvantages, and the application needs to choose the most suitable algorithm according to the specific needs and environmental conditions. With the improvement of computing power and optimization of algorithms, dynamic feature detection technology is expected to show more potential and value in many fields in the future [5].

*2.3. Image Acquisition of Floating Objects in River and Lake Based on AI Intelligent Image Recognition*

In the realm of intelligent image recognition, particularly in the context of monitoring river and lake environments, the identification and tracking of floating entities within water bodies, such as vessels, aquatic flora, and pollutants, have emerged as pivotal areas of research. The fundamental objective of this technology lies in the efficient acquisition and analysis of imagery to autonomously detect and identify various types of floating objects within specific water bodies, thereby enabling the assessment of water quality or the provision of necessary environmental protection measures [6].

Taking the example of a water intake area in a reservoir (as depicted in Figure 2), this zone is frequently traversed by vessels and is also prone to the presence of various naturally occurring or deliberately introduced floating pollutants, including but not limited to water hyacinths, cyanobacteria, and drifting debris. These floating entities not only compromise water quality but also pose potential threats to the health of underwater ecosystems and the safety of waterway utilization.

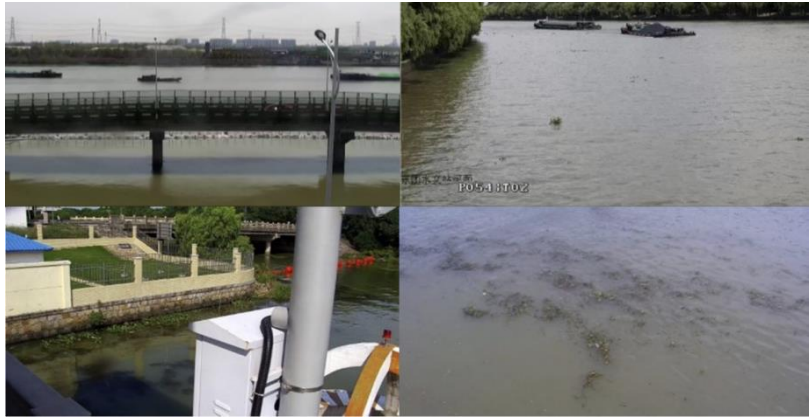

**Figure 2.** Main research objects

Utilizing artificial intelligence (AI) technologies, particularly deep learning and computer vision algorithms, facilitates the continuous surveillance of these aquatic environments. By deploying cameras around the water bodies or utilizing imagery collected by drones, AI models can autonomously identify floating objects within the images. This process entails image preprocessing, feature extraction, and classification steps, ensuring precise and efficient detection of floating debris [7].

Through this sequence of operations, AI-powered image recognition systems proficiently discern target objects from intricate aquatic surroundings, such as vessels, water hyacinths, and cyanobacteria, subsequently categorizing and pinpointing them. This not only extends technical assistance to water quality monitoring but also furnishes environmental protection and management authorities with timely and precise data support, facilitating the implementation of pertinent conservation and remediation measures.

**3. Deep learning based target detection algorithm for floating objects in rivers and lakes**

*3.1. Deep learning based target detection framework*

In the exploration of detecting floating objects in rivers and lakes, deep learning techniques, particularly Convolutional Neural Networks (CNNs), play a pivotal role. This technology, by analyzing and processing vast amounts of image data, autonomously learns and extracts crucial features from images, thereby achieving precise identification and classification of floating objects. Deep learning frameworks, through the continual optimization of weights during training, adeptly capture image features, thereby enhancing the performance and accuracy of detection algorithms.

CNN architectures such as LeNet-5, ResNet, and MobileNet have demonstrated their superior performance on international platforms like the ImageNet Large Scale Visual Recognition Challenge (ILSVRC). These models, ingeniously designed, can handle image recognition tasks ranging from simple to complex, with their primary distinctions lying in the depth, breadth, and innovative details of network structures [8].

LeNet-5, as an early CNN model, was primarily utilized for handwritten digit recognition, laying the groundwork for the subsequent development of deep learning models. The LeNet-5 model is illustrated in Figure 3.

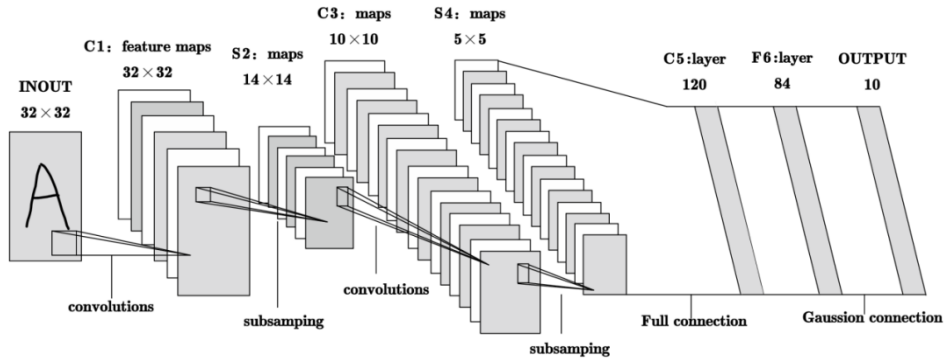

**Figure 3.** LeNet-5 model

ResNet, by introducing a residual learning framework, addresses the degradation issue in deep network training, effectively augmenting the depth of networks and enhancing recognition accuracy.

On the other hand, MobileNet targets applications on mobile and embedded devices, achieving lightweight network architectures through depth-wise separable convolutions, maintaining high recognition performance while reducing computational overhead and model size.

The application of these deep learning frameworks, particularly in the domain of floating object detection in rivers and lakes, enables efficient and precise identification of various floating objects such as boats, aquatic plants, and various pollutants. This holds significant practical significance for water quality monitoring, environmental protection, and water resource management. In the future, with technological advancements, these deep learning models are poised to achieve even greater improvements in accuracy, speed, and applicability [9].

*3.2. SSD-based target detection of floating objects in rivers and lakes*

In the realm of river and lake floating object target detection, the adoption of the Single Shot MultiBox Detection (SSD) algorithm for investigation signifies an exploration into efficient approaches to tackle this issue. The crux of the SSD algorithm lies in its ability to swiftly and accurately identify and localize the size and shape of floating objects through a distinctive pyramid-style feature map structure. This method leverages feature maps of different resolutions to effectively detect targets of various sizes, while combining predefined bounding boxes (prior boxes) significantly enhances the speed and precision of detection.

The innovation of the SSD algorithm lies in its amalgamation of classification and localization into a unified detection mechanism. Through a single-step processing, it directly predicts the category and position of objects, circumventing the complexity of multi-step processing in traditional detection systems. This one-step detection approach not only simplifies the model structure but also reduces the difficulty of model training, while ensuring the efficiency of the detection process. The principles of SSD river and lake floating object target detection are depicted in Figure 4.

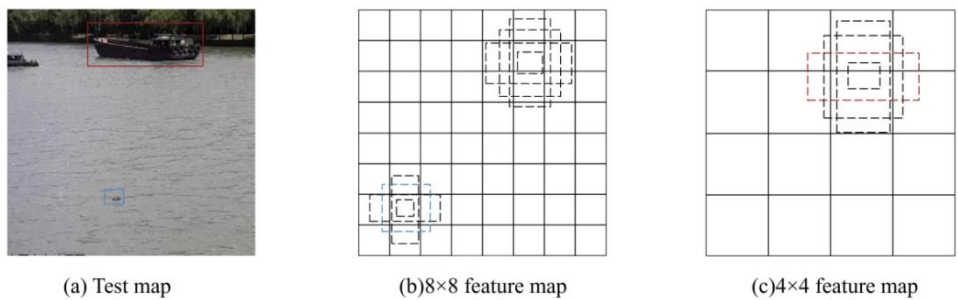

**Figure 4.** Principle of SSD-based floating object detection in rivers and lakes

By implementing SSD-based algorithms for river and lake floating object detection, it becomes feasible to achieve swift and real-time monitoring of surface debris. This holds significant practical value in promptly identifying and addressing sources of pollution in water bodies, thereby upholding the environmental safety of aquatic ecosystems. With the continuous advancement of deep learning technologies, SSD-based models for river and lake floating object detection are poised to become even more precise and efficient, offering robust technical support for the monitoring and preservation of aquatic environments.

*3.3. Faster-RCNN-based target detection of floating objects in rivers and lakes*

In the realm of research where deep learning techniques are applied to the detection of floating objects in rivers and lakes, the Single Shot Multibox Detector (SSD) algorithm has demonstrated its remarkable contributions to both processing speed and accuracy. In contrast to traditional two-stage detection algorithms such as the Faster R-CNN and other R-CNN variants, SSD adopts an innovative one-step implementation strategy in its design, aiming to accomplish object recognition and localization simultaneously through a single inference process. This algorithm addresses the challenge of detecting floating objects of various sizes and shapes effectively by constructing a series of feature maps at different scales and utilizing different-sized prior boxes on these feature maps to predict the positions and categories of objects. The core advantage of SSD lies in its ability to leverage the rich features extracted by deep convolutional networks without the need for the complex process of generating candidate regions followed by classification and localization, thus significantly improving detection speed. This rapid and direct detection approach not only reduces the complexity of model training but also enhances the algorithm's adaptability to dynamic environmental changes in practical applications, rendering it highly valuable in river and lake environment monitoring, water quality management, and waterway security. By continuously optimizing the performance of the SSD algorithm, it is expected to achieve more precise and efficient automatic detection and identification of floating objects in rivers and lakes in the future [10].

*3.4. Research on floating object detection in river and lake based on YOLOv5*

In the evolution of technology for detecting objects on water bodies, the YOLOv5 algorithm emerges as an advanced paradigm in deep learning, serving as a pivotal instrument in addressing real-time detection challenges. Leveraging its distinctive architectural design, YOLOv5 swiftly discerns and locates objects within images. Its operational framework revolves around a singular image analysis process, enabling rapid identification of multiple objects within an image, accompanied by precise classification and positional data. Notably, YOLOv5 stands out for its optimized feature extraction and processing pipeline, bolstered by the integration of deeper yet more efficient convolutional networks, thereby enhancing the model's capacity to detect small-sized and blurred objects. Furthermore, the introduction of multiscale training during the training phase endows the model with superior adaptability and robustness to objects of varying sizes. These optimizations ensure YOLOv5's efficacy and high precision in the task of detecting floating debris in rivers and lakes, particularly in scenarios necessitating prompt responsiveness and management of large-scale environmental data, underscoring its remarkable potential for application.

## 4. River and lake floating object detection system design

*4.1. Floating Object Detection System Hardware Platform Construction*

In the illustration, Figure 5 depicts a platform for detecting floating objects on the water surface, which relies on a sophisticated camera capable of capturing high-definition images of the water surface and a powerful server. Specifically, the system utilizes a high-resolution camera designed to capture real-time video streams of rivers or lakes, ensuring the accuracy and efficiency of data acquisition. Subsequently, this video data is transmitted via Real-Time Streaming Protocol (RTSP) to a high-performance processing server, selected for its robust processing capabilities. Its primary task is to receive, analyze, and execute complex algorithms for detecting floating objects from the camera feed. Upon completion of the analysis and identification process, a server built using the Flask framework is responsible for providing real-time feedback of the processing results to the user. This design not only optimizes the data processing workflow, ensuring the real-time and precise nature of the detection task but also greatly enhances the overall performance and applicability of the river and lake floating object monitoring system through the integration of advanced hardware platforms and software technologies.

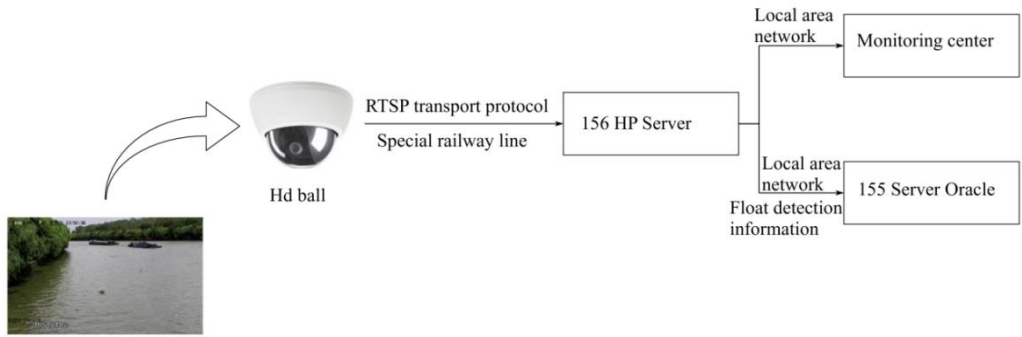

**Figure 5.** Floater target detection platform

*4.2. System software framework design*

The software architecture of the river and lake floating object detection system is divided into two main levels: the application layer and the middleware layer. The application layer is responsible for implementing intuitive user interaction and core functionalities, including real-time monitoring of video streams, automatic identification of floating objects, and graphical presentation of results. This layer directly influences the user's operational experience and the practicality of the system. Meanwhile, the middleware layer serves as a bridge, processing video streams captured by cameras to the computer system via the RTSP protocol and facilitating efficient inter-process communication among different modules. Such layered design not only clearly delineates the functional areas of the system but also optimizes the data processing workflow, ensuring that each step from video capture to floating object detection and final presentation runs efficiently and stably.

## 5. Experimental results analysis

*5.1. Model hyperparameter setting and training*

Target detection models in a deep learning framework rely on large-scale parameter tuning and weight optimization, a process that is extremely demanding on computational resources. In order to train such networks efficiently, processed datasets with data augmentation techniques are employed with the aim of improving the generalization ability and accuracy of the models by increasing the diversity of the data. To train such models, it is necessary to equip hardware with powerful computational capabilities and a suitable software environment. Such a configuration ensures high efficiency and stability when dealing with complex algorithms and huge datasets, which enables the model to learn enough features to improve the accuracy of detecting floating objects in rivers and lakes, and the system's hard- and software-configured environments are shown in Table 1.

**Table 1.** System hardware and software configuration environment

| Disposition | Argument |
|---|---|

| CPU | CPU: Intel(R)Xeon(R)Silver 4110 CPU @ 2.10GHz 2.10GHz (2 processors) |
|---|---|
| GPU | RTX 2060 SUPER |
| Internal memory | 128GB |
| Video memory | 8GB |
| Software environment | Windows10 Pro, Python3.8, Pytorch 1.8.1, CUDA10.2 |

*5.2. Test results and comparative analysis*

Three different deep learning network models, Faster-RCNN, SSD, and YOLOv5, were used for the analysis in the river and lake floating object detection study. These models have undergone targeted training using datasets derived from those described in Chapter 2 and constructed specifically for water surface floating object detection.The SSD model is effective in identifying vessels on the water surface, although it exhibits limitations in capturing smaller objects such as water hyacinths or small vessels in the distance, leading to an increase in the leakage rate.The detection results of the SSD algorithm are plotted in Figure 6.The target detection of this model has a confidence is low, although its classification accuracy is fair.

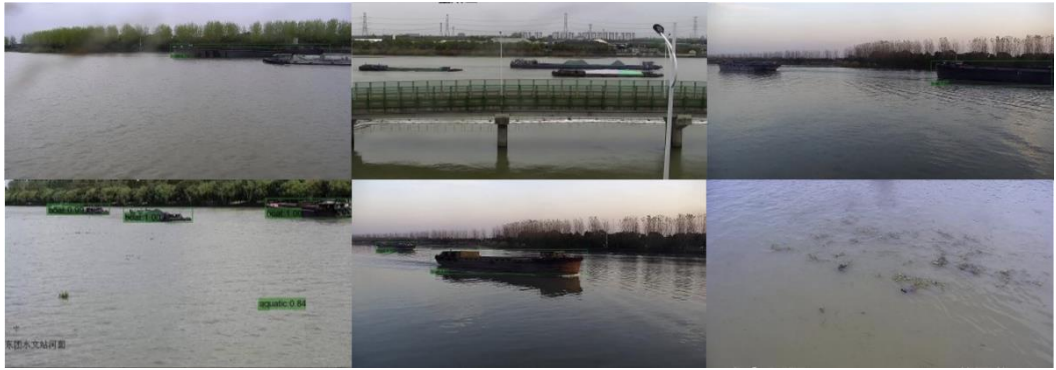

**Figure 6.** graph of detection results of SSD algorithm

The Faster-RCNN model exhibits higher performance, especially in dealing with scenes where targets are occluded from each other, and it can provide more accurate target localization with a confidence range usually between 0.7 and 0.9, Figure 7 shows the Faster-RCNN detection results. However, it also faces some misdetection and omission problems due to small target sizes or overly complex backgrounds, such as distant bridges or railings.

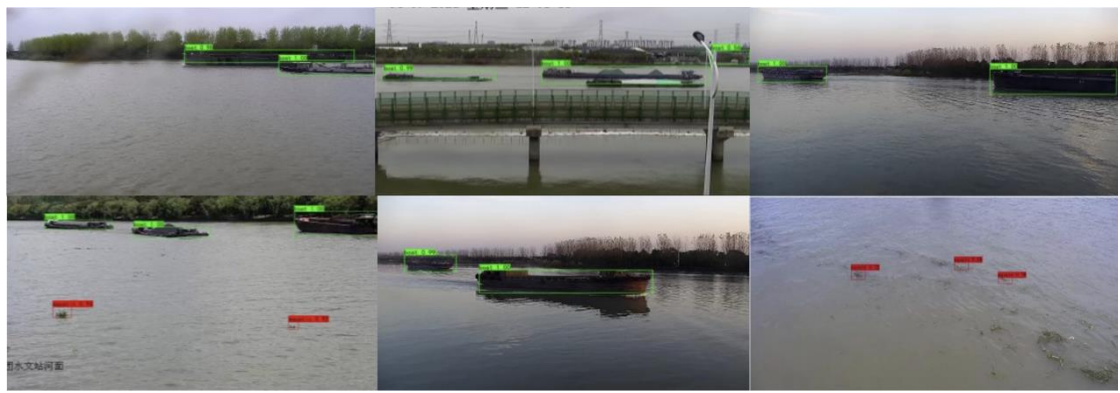

**Figure 7.** Graph of Faster-RCNN detection results

In most cases, the YOLOv5 model is able to recognize the target, even in the case of target occlusion or insufficient light.Figure 8 shows the effect of YOLOv5 detection, which is slightly less accurate than Faster-RCNN, but shows high accuracy in target localization and classification.

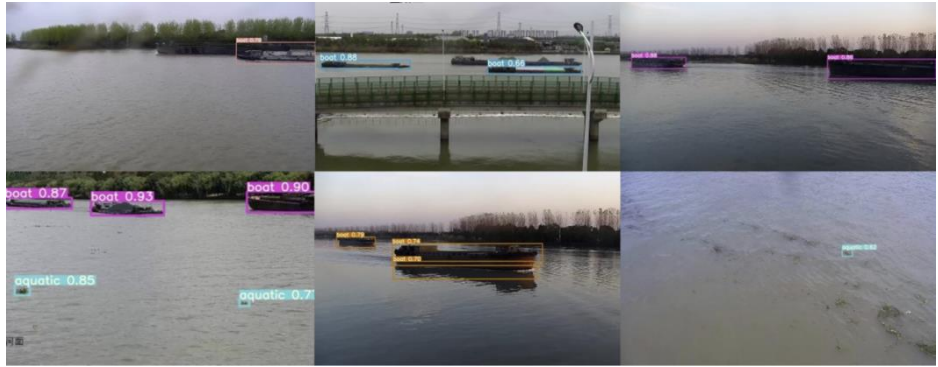

**Figure 8.** Effectiveness of YOLOv5 Detection

Taken together, each of these three models has its own advantages and limitations, and all of them have certain application value and effect in the task of detecting floating objects in rivers and lakes, reflecting the potentials and challenges of deep learning technology in practical applications.

## 6. Conclusion

Through the depth and exploration of this study, the river and lake floating debris detection technology based on AI intelligent image recognition has demonstrated its strong application potential and practical value. It not only realizes the innovation in technology, but also proves its significant contribution to improve the efficiency and accuracy of water quality monitoring in practice. Facing the global challenge of environmental protection, technological innovation provides us with new perspectives and methods. In the future, with the continuous progress and optimization of AI technology, it is believed that the detection and management of river and lake floaters will become more intelligent and accurate. This is not only a great contribution to the field of environmental science, but also a solid step towards a greener and more sustainable future.